# A Real-time Face Mask Detection and Social Distancing System for COVID-19 using Attention-InceptionV3 Model

*Abdullah Al Asif[1,*], Farhana Chowdhury Tisha[2]*

[1]BRAC James P Grant School of Public Health, BRAC University, Dhaka-1213, Bangladesh
[2]Department of Computer Science and Engineering, East West University, Dhaka-1212, Bangladesh

**ABSTRACT**

One of the deadliest pandemics is now happening in the current world due to COVID-19. This contagious virus is spreading like wildfire around the whole world. To minimize the spreading of this virus, World Health Organization (WHO) has made protocols mandatory for wearing face masks and maintaining 6 feet physical distance. In this paper, we have developed a system that can detect the proper maintenance of that distance and people are properly using masks or not. We have used the customized attention-inceptionv3 model in this system for the identification of those two components. We have used two different datasets along with 10,800 images including both with and without Face Mask images. The training accuracy has been achieved 98% and validation accuracy 99.5%. The system can conduct a precision value of around 98.2% and the frame rate per second (FPS) was 25.0. So, with this system, we can identify high-risk areas with the highest possibility of the virus spreading zone. This may help authorities to take necessary steps to locate those risky areas and alert the local people to ensure proper precautions in no time.

## 1 Introduction

In December 2019, Wuhan, the People's Republic of China, reported the discovery of a new virus called 2019-nCoV. The virus became an outbreak in 2020 around the world. It has caused the 21st century's global health crisis. The condition has been declared a COVID-19 pandemic by the World Health Organization (WHO). According to the world meters data, more than 250 million people have been impacted, with 5.1 million individuals dying around the world [1]. This virus is growing extremely powerful as it develops the capacity to modify its pattern on its own, making it increasingly difficult to erase. As this virus is highly infectious, it can infect a substantial percentage of people within a short period. Fever, cough, runny nose, chills, weakness, body ache, loss of taste and smell, and shortness of breath are all usual symptoms. Since most of these symptoms are identical to the influenza virus, many individuals ignore them until they become severe, which in some cases, death results. This severity has compelled governments in several nations to implement lockdowns to prevent the virus from spreading, resulting in a massive crisis in many industries. Scientists and the World Health Organization (WHO) confirmed that this virus transmits through human interaction, so they have recommended some protocols for safety measurements to reduce the transmission of COVID-19. By wearing a facemask and maintaining social distancing, people can ensure their safety in some situations [2]. An infected person should not interact with anyone and should stay at home if they do not need a medical emergency. Wearing a facemask and keeping social distance are also effective ways to reduce physical contact. As a result, our primary goal in the current context is to give technical services to make sure that these two indicators are appropriately enforced. Some dedicated research works have been done for facial recognition or measuring distance calculation, but we never expected that facemask detection or ensuring maintenance of social distancing could be a vital aspect considering the current situation. For this purpose, many researchers mostly considered various versions of the YOLO algorithm for implementing those safety measurements. But most of them faced the scarcity of datasets, not enough research materials on this area, not enough implementation of different types of models for the comparison purpose as well. Our objective is to find out that the two of the basic safety measurements of COVID-19 such as wearing facemask and maintenance of social distancing are properly getting maintained with having minimal infringements by people in public places.

We used deep learning approaches to identify facial masks and social distance between persons. As we know that the attention model works best for identifying any particular object on the real-time dataset. Besides, the inceptionv3 model also works well for object detection purposes as it has the advantages of extra layers in it with minimal computational cost in the shortest time. That is why we combined these two models in our system. We have used inceptionv3 for pre-train purposes and the attention model as the final layer in it. We have summarized them in the points below:

In both situations, we employed the object detection approach and the method is capable of detecting anomalies in real-time video feeds.

## 2 Related Works

Almost the whole world got to face the outburst of the COVID-19 tragedy. This infectious disease shows almost similar symptoms to the influenza virus, but still, medical researchers and pharmaceuticals are unable to invent the appropriate remedy for this virus. It is very difficult to say how long it will take to find the proper medication or vaccination for the disease. According to WHO, reduction in human physical interaction can decrease the transference of COVID-19, so that is why considering the current situation, it has become the most significant matter to contemplate. Many researchers around the world also tried to use some tracking devices or drones using Bluetooth and GPS to locate covid affected people and in India, Aarogya Setu was one of them [3]. The application was intended to be made for Indian people mostly as this app supported many Indian languages. Some of them used wireless signal amplitude wave-form for human detection as well [4]. But in this case, a huge number of antennas needed to be implemented to work out this properly, but this might cost a lot of time and money. As the Covid-19 outburst is very recent, so there are not so many papers that have been researched on this issue particularly. But Face masks and human detection can be considered object detection in computer vision. However, object detection and classification are very well-known terms in the technical world. In state-of-the-art, the advancement of computer vision and deep learning methods has been shown immense progress in so many circumstances. There are lots of popular models and algorithms like Convolutional Neural Network (CNN), Recurrent Neural Network (RNN), etc. in deep learning that has been used in this subject. Convolutional Neural Network is one of the deep learning categories that has been accomplished great performing capability in image recognition standards. Different CNN models such as AlexNet, InceptionV3, ResNet-50, VGG16, etc. are showing great success in detecting objects in various scenarios due to high-performance computing systems [5]. CNN is also undoubtedly used for the classification part and for getting more information from an image.

---

*Corresponding Author Email Address: asifabdullah6@gmail.com

Many researchers mostly acknowledged one of the CNN-based model's various versions of the YOLO (You Only Look Once) algorithm for optimized object detection in real-time [6]-[9]. YOLOv3 algorithm has been considered for detecting social distance by using the 2-dimensional top-down view of video footage where the camera angle was fixed [10]. They have monitored and calculated the distance between pedestrians by creating a fixed square-shaped place on the street. Their model can also detect the pedestrian's half body when they enter the box. But it cannot show any result until the object is inside of the box. So, if any object avoids that fixed place, the system will not be able to detect it. If this limitation is revealed, people easily would be able to deceive this system.

Another researcher proposed a model which is a combination with an inverse perspective mapping (IPM) technique with the YOLOv4-based Deep Neural Network (DNN) model, and it is applicable for different environments using CCTV cameras [6]. For face detection in real-time, the Haar Cascade classifier and YOLOv3 algorithm have been used as well [8]. A device like glasses was made for blind people with an alarm to maintain social distancing for the Covid-19 issue [11]. But the device does not warn the person of anything except other persons is in the reported distance. These devices were based on GPS, Wi-Fi, and networks and that is why those devices cannot work when the device will lose every type of connection from it. Some of the other developers used two other CNN-based models in their hardware devices such as VGG16, MobileNetv2 as well [12],[13].

Face detection was widely used in the state-of-the-art but detection of facemask for limiting the transference of the virus is now a very important event in the current situation. People need to ensure that they are properly wearing their facemask and keep following social distancing as very basic steps in public places.

Another noticeable fact was that there was no implementation of these preliminary steps together in one system. After observing these research gaps, we have proposed a model with the maintenance of those two safety measures which were not supposedly used for this purpose more accurately in the shortest time till now.

## 3 Methodology

The suggested model has two primary components: The model is deep transfer learning for face mask identification with the customized inception-v3 model, and the second one is calculating the Euclidean distance between pixels of an image. The Inception network has been meticulously designed. Google's Inception V3 is the third version in a series of Deep Learning Convolutional Architectures. Inception V3 was trained with 1,000 classes from the original ImageNet dataset, which was trained with over 1 million training pictures [14].

In our research findings, we did not notice that any other researcher used this model for this purpose yet. We want to build this system for real-time, so time efficiency is very much important. In that case, the attention mechanism plays a vital role. Generally, the concept of the attention model is inspired by the function of the human eye. Our eyes can catch a fairly wide perspective of the world, but we tend to "glance" at the overview and only pay attention to a small portion of it, leaving the rest "blurred out." The attention model tries to capitalize on this concept by allowing the neural network to "focus" its "attention" on the most interesting part of the image, where it can get the most information while paying less "attention" elsewhere, resulting in a reduction in the amount of image processing power required [15]. Besides, the inceptionv3 model also works well for object detection purposes as it has the advantages of extra layers in it. That is why we combined these two models in our system. We have used inceptionv3 for pre-train purposes and the attention model as the final layer in it.

### 3.1 Data Collection

The experiments in this study were carried out on two different datasets along with 10,800 images including both with and without Face Mask images. The first dataset has taken from a website (https://makeml.app/datasets/mask) under a public domain license with 1776 images [16]. The rest of the 9024 images were collected and filtered from the search engines using Bing-image-downloader [17]. This is known as a python library to download the bulk of images from Bing.com. For the training, validation, and testing phases, both datasets were used.

### 3.2 Facemask Detection

The purpose is to create a facemask detection system by retraining certain modified final layers with attention mechanisms and utilizing a pre-trained inceptionv3 as a base model. So, this part consists of two parts which are feature extraction and classification. The pre-trained inceptionv3 model is deployed as a standalone feature extraction part, in which case input images are pre-processed by the model or a portion of the model to produce an output (e.g., a vector of numbers) that may subsequently be utilized as input for training a new model. The default input image size of Inception-v3 is 299×299; however, the image shape resized into 150×150 pixels in this model. This did not affect the number of channels, but only on the size of the feature maps created during the method, and the outcome was adequate. The feature map dimensions after the convolutional layers and Inception modules were 3×3 with 2,048 channels as in shown Fig. 1. This is all about the extraction of general features from input images.

Now the visual soft attention mechanism comes to play to classify the images based on the features obtained from the inceptionv3 model. The attention model is a streamlined attempt that focuses on the few relevant things of the selective activities [18]. As a result, this classification system focuses on human faces and determines whether or not the individual is wearing a mask.

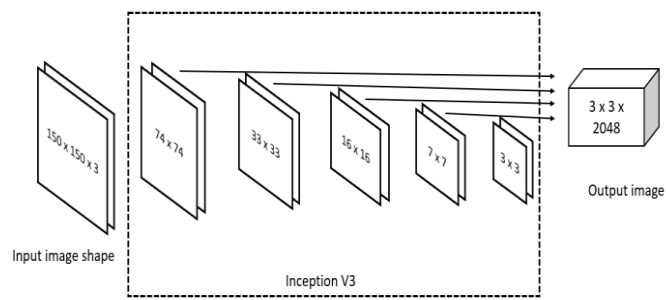

Fig. 1 The model architecture of inceptionV3

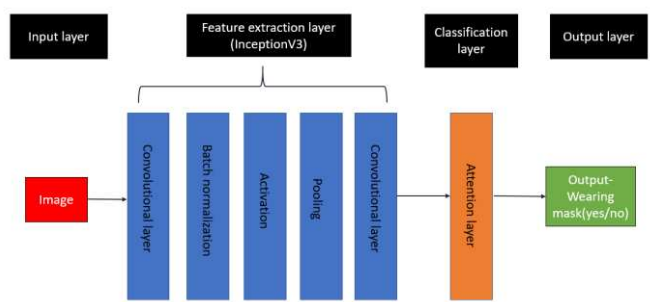

Fig. 2 The proposed transfer learning model

In transfer learning, the method reuses the feature extraction component and retrains the classification section with the desired dataset. As the feature extraction section (which is the most complex part of the model) does not have to be trained, as shown in Fig. 2, it takes fewer computer resources and less training time to train the model.

### 3.1. Measuring Social Distance (SD)

People cannot share germs if they are not close together, hence social distance is undoubtedly the most effective prevention technique to limit disease spread. This section focuses on determining the distance between persons in public places using our suggested technique. The human face recognition and tracking module receives video stream sequences from the cameras. To measure the level of social distance practiced, factors such as the center of a person's position and distance between them play a vital role [19]. The color of the bounding box, green to red, will be changed by an alert for the social distancing violation.

First, we locate any human face and denote it as the marker. Then, we calculate its width in pixels by taking the difference of the horizontal coordinates of the bounding box's top-left corner ($X_1$) and bottom-right corner ($X_2$) and denote it as $P = X_2 - X_1$ [20].

The distance between two people determines how close they are. The decision is based on a comparison of the distance vector with a threshold value. If the Euclidean distance between two things is less than a certain threshold value, it is thought that they are not adhering to the social distancing rules or have not created adequate distance between them. The distance between two pixels is calculated by the following formula:

Euclidean distance, $d(p,q) = \sqrt{\sum_{i=1}^{n}(q_i - p_i)^2}$,

where, p, q = two points in Euclidean n-space, n = Number of dimensions, $q_i$, $p_i$ = Euclidean vectors, starting from the origin of the space (initial point).

We applied a 2D Euclidean distance system measurement in this case. The face will be detected using the distance measurement function, which will return the face width as a pixel value. The focal length of the camera is calculated using the pixel value. Finally, we calculated the distance using a reference image. To measure distance, we compare the pixel of the reference picture to the pixel of our acquired image. We need to know the exact size of an item in the image to estimate distance on a real scale. That is why the reference image is being used. If we know the distance between the ground level and the camera on a big scale, we can simply determine the distance. However, if we don't know the reference value, we can't compute the distance using the 2D technique. This is the system's most serious flaw. As a result, we intend to apply the 3D reconstruction approach in our future work.

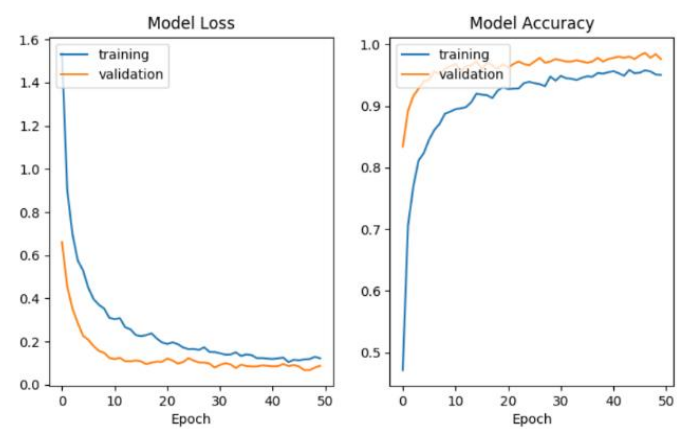

Fig. 3 Model training accuracy and loss curve

## 4  Experimental Results and Discussion

This paper tries to fine-tune using Inception V3 on a dataset containing around ten thousand images in two classes. Initially, we froze all base layers and just trained 2 fully connected layers (2048 units and 3 respectively). The training accuracy went 98 percent in 20 epochs and the validation accuracy is 99.5 percent, as shown in Fig. 3. The model initially detects all people in the cameras and shows a green boundary case around each person who is far from the other. If the minimum distance (threshold) exceeds, which means people violate social distance (SD) norms the color of the box changes to red [21].

In Fig. 4, the child is properly wearing a mask, but still, the bounding box is red because our model not only detects facemask but also measures the social distancing aspects as well and as we can see that part did not maintain properly so that is why the bounding box turned into a red one.

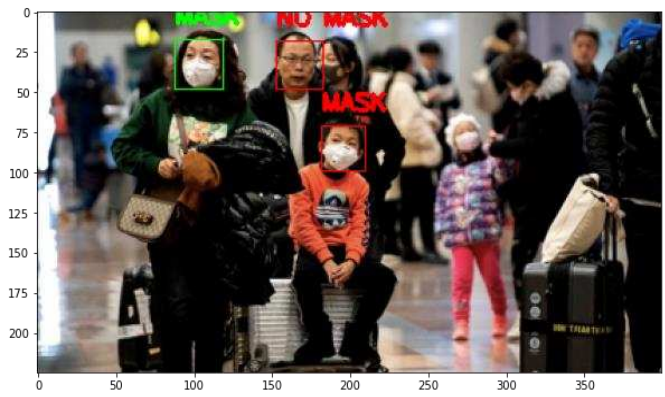

Fig. 4 Visualization of the test results obtained by the proposed system

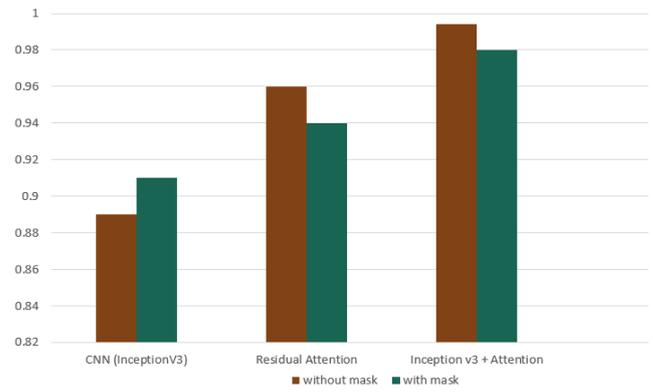

Fig. 5 The f1-score for CNN, attention mechanism, and hybrid inceptionV3+attention model

The model detects the SD and masks with a precision score of 98.2 percent and a recall value of 97.8. The frame rate per second (FPS) was 25.0.

The experiments were conducted on Google Colaboratory with Intel(R) core i5, 1.80 GHz CPU, Virtual TPUv2-8 64 GB, 12GB DDR4 VRAM, and 8 GB RAM. All programs were written in Python - 3.6 and utilized OpenCV - 4.2.0, Keras -2.3.0 and TensorFlow - 2.2.0. We have mentioned different types of models and their accuracy which have been used relatively in this sector in Table 1.

Table 1 Comparison table of other models with details

| References | Algorithm | Accuracy | Data |
|---|---|---|---|
| [5] | YOLOv4 | 99.8 | 3,762,615 |
| [6] | YOLOv2 | D1-95.6% D2-94.5% | 1575 (d1-775, d2-800) |
| [9] | VGG-16 | 96% | 25000 |
| **This paper** | **InceptionV3+ attention** | **99% to 99.5 %** | **10800** |

As we can see many kinds of methodology considered only one safety measurement which was mask detection but our model ensures two of them in the meantime. Most of them have been chosen to go with the YOLO algorithm and its different versions. After comparing the details, we have noticed that another type of the proposed model had obtained 99.8% accuracy with almost 3.5 lakhs dataset, but our dataset only consists of 10800 data, and as we know the inceptionv3 model provides higher accuracy with small datasets [21]. Since the system initially would not have enough data but as it will work in a real-time environment so after the implementation it would be able to collect enormous data. Additionally, the attention mechanism is known for identifying individual small objects more precisely from an image, and as our system needed to find facemasks from a real-time scenario so that is why we had chosen this model for our research purpose and successfully got 99.5% accuracy.

The test accuracy for different algorithms is shown in Fig. 5 and shows the performance of the f1-score for with mask and without mask classification.

## 5 Conclusion and Future work

The spread of the contagious COVID-19 virus has created global health crises all over the world, so wearing facemasks and maintaining social distancing are the preliminary precautions. Our model works in real-time so that it has the potential to reduce violations of those two criteria and it can improve the safety of people in the meantime which can reduce the spreading rate of COVID19 all around the globe. By discovering those people who are not wearing masks and not ensuring social distancing, we can ensure that they do so and that is how we can make sure public safety.

We can implement this model in a hardware device for monitoring where the COVID-19 affected people are living the most so that we can ensure that those people are strictly maintaining face masks and social distancing. We can also update another feature where it will detect whether a person is wearing double facemasks, face shields, or not. Besides, one of the common symptoms of COVID-19 is fever and, in our system, we can implement a thermal camera through which we can measure the temperature of a person from a very far distance. We can also use an approach where it will be able to detect if a person is having a breathing problem or not by observing the person's activity. This system cannot only be used for COVID-19 but also it can be used for any kind of infectious disease as well.